\documentclass[10pt,twocolumn,letterpaper]{article}

\usepackage{wacv}
\usepackage{times}
\usepackage{epsfig}
\usepackage{graphicx}
\usepackage{amsmath}
\usepackage{amssymb}

\DeclareMathOperator*{\argmax}{arg\,max}
\DeclareMathOperator*{\argmin}{arg\,min}
\usepackage[ruled,vlined]{algorithm2e}
\usepackage{caption} 
\usepackage{multirow}

\usepackage{pifont}
\newcommand{\cmark}{\ding{51}}
\newcommand{\xmark}{\ding{55}}

\usepackage[numbers,sort]{natbib}

\def\wacvPaperID{1127} 

\wacvfinalcopy 

\ifwacvfinal
\def\assignedStartPage{1} 
\fi

\ifwacvfinal
\usepackage[breaklinks=true,bookmarks=false]{hyperref}
\else
\usepackage[pagebackref=true,breaklinks=true,colorlinks,bookmarks=false]{hyperref}
\fi

\ifwacvfinal
\else
\fi

\begin{document}

\title{Few-Shot Learning with No Labels}

\author{Aditya Bharti\\
{\tt\small aditya.bharti@research.iiit.ac.in}
\and
Vineeth N. B.\\
{\tt\small vineethnb@cse.iith.ac.in}
\and
C. V. Jawahar\\
{\tt \small jawahar@iiit.ac.in}
}

\maketitle

\begin{abstract}
Few-shot learners aim to recognize new categories given only a small number of training samples. The core challenge is to avoid overfitting to the limited data while ensuring good generalization to novel classes. Existing literature makes use of vast amounts of annotated data by simply shifting the label requirement from novel classes to base classes. Since data annotation is time-consuming and costly, reducing the label requirement even further is an important goal. To that end, our paper presents a more challenging few-shot setting where no label access is allowed during training or testing. By leveraging self-supervision for learning image representations and image similarity for classification at test time, we achieve competitive baselines while using \textbf{zero} labels, which is at least \textbf{10,000 times} fewer labels than state-of-the-art. We hope that this work is a step towards developing few-shot learning methods which do not depend on annotated data at all. Our code will be publicly released at \url{https://anonymous}.
\end{abstract}


\section{Introduction}
\vspace{-3pt}
\begin{figure}
\centering
\includegraphics[width=\linewidth]{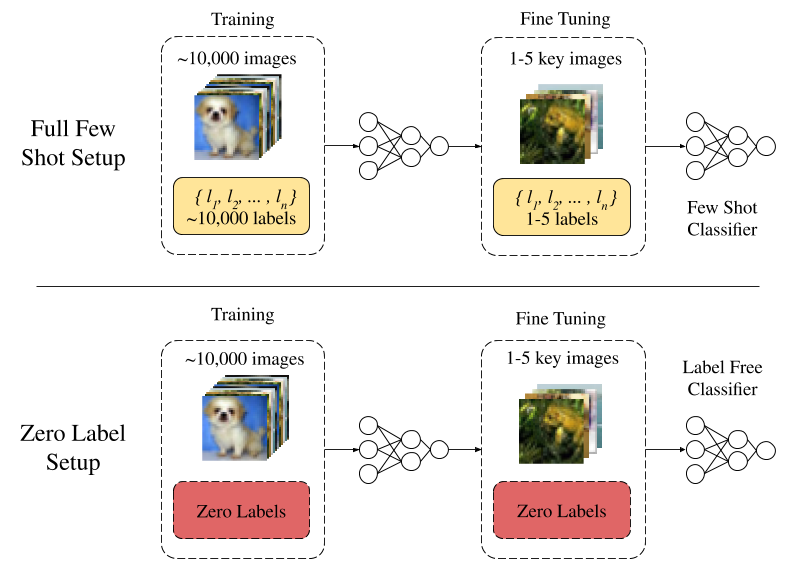}
\vspace{-2pt}
\caption{Few-shot image classification without training or testing labels. By leveraging self-supervision for training and image similarity for testing, we achieve competitive baselines without using any labels at all.}
\vspace{-5pt}
\label{fig:example_test}
\end{figure}

Few-shot learners~\cite{matchNets, protoNets, pmlr-v70-finn17a} aim to learn novel categories from a small number of examples. Since getting annotated data is extremely difficult for a large number of natural and man-made visual classes~\cite{2018fine_grained_wild}, such systems are of immense importance as they alleviate the need for labeled data.
Few-shot learning literature at this time is  diverse~\cite{wang2019generalizing} with multiple categories of approaches. Meta-learning~\cite{pmlr-v70-finn17a,Ravi2017OptimizationAA,tadam} is a popular class of methods which use experience from multiple base tasks to learn a base learner which can quickly adapt to novel classes from few examples. There has been immense progress using the meta-learning frameworks over the last few years~\cite{prob_MAML, bayes_MAML, recast_MAML, recast_MAML, vae_MAML, how_MAML}. While extremely popular, such approaches are computationally expensive, require that the base tasks be related to the final task, and need a \textit{large number of training labels} for the base tasks. Other approaches focus on combining supervised and unsupervised pipelines~\cite{gidaris_ssl_boost,maji_ssl_boost,amdim_net} and others alleviate the data requirement by generating new labeled data using hallucinations~\cite{hallucinate_hariharan}. Recent methods~\cite{wang2019simpleshot, chen2020simple} have also established strong baselines with computationally simple training pipelines and classifiers. However, all these methods use label information at some stage or the other. In our work, we directly address the label requirement in few-shot learning by developing a completely self-supervised few-shot learning method with simple components that learn and classify images with no labels at all.

Recent work in contrastive learning~\cite{ji2019iic, he2019moco, chen2020simple} has shown that it is possible to learn useful visual representations without labels by learning image similarity over multiple augmented views of the same data, paired with a suitable training strategy and a loss function. We leverage SimCLR~\cite{chen2020simple} and MoCo~\cite{he2019moco} to develop label-free training methods for few-shot learning. Since image similarity is an effective pre-training task for few-shot~\cite{Koch2015SiameseNN}, we perform image classification using image similarity as shown in Figure \ref{fig:example_test}. Instead of predicting image labels, our classifiers choose a key image which is most similar to the input to be classified. This allows us to classify images without using labels at test time.

The key contributions of our work can be summarized as follows:
\begin{itemize}
\vspace{-3pt}
\setlength\itemsep{-0.1em}
  \item We present a new, challenging label-free few-shot learning methodology.
  \item Our few-shot learning methodology is both computationally and conceptually simple in terms of training methods and pipeline components, making it easy to adopt or adapt.
  \item By leveraging self-supervision for training and image similarity for testing, we achieve competitive performance while using \textit{zero} training or testing labels. This is \textit{10,000 times} fewer labels than existing state-of-the-art.
  \item We also report studies on use of clustering methods in our pipeline, as well as the use of labels when available inside our methodology, and show how this affects/improves performance.
\end{itemize}

\section{Related Work}
\vspace{-3pt}
In this section, we discuss few-shot learning methods in general, followed by a more detailed discussion on more recent approaches, and then present efforts from related fields which are connected to our work.

\subsection{Few Shot Learning}
\vspace{-3pt}
The core challenge in few-shot learning is the lack of sufficient supervisory signals on novel classes to effectively learn a classifier. One category of methods seeks to augment the training set by using prior knowledge, and then train their algorithms on the larger dataset. Methods such as \cite{hallucinate_hariharan, schwartz2018delta, kwitt2016one} use the existing training set to generate new samples by modeling the intra- and inter-class variations. In cases where a separate weakly supervised or unlabeled dataset is present, approaches such as \cite{douze2018low, pfister2014domain} choose informative examples and propagate labels from the existing training set to label the new examples. Finally, approaches such as \cite{tsai2017improving, gao2018low} use completely different datasets as extra information to guide the learning process. However, these approaches require the presence of related datasets, or a reliable method to select informative examples, or the ability to correctly propagate labels to new examples. In contast to all thse methods, our approach does not use labels at all.

\begin{figure*}
  \centering
  \includegraphics[width=\linewidth]{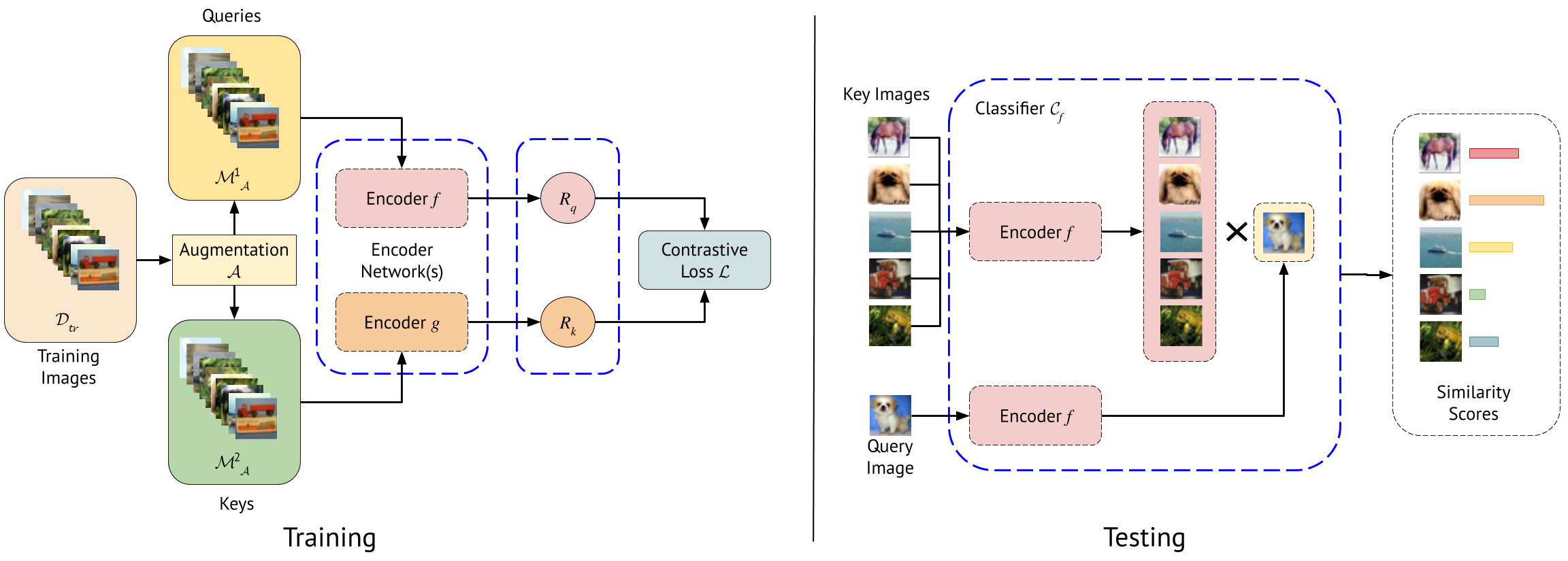}
  \vspace{-12pt}
  \caption{General overview of our training and testing methods. \textbf{Left}: Self-supervised training to learn contrastive representations without labels. An input minibatch is passed through an augmentation module, which generates key and query image batches using image transforms. Encoder networks then learn to represent these images so as to preserve image similarity between images generated from the same input. The overall pipeline is trained using contrastive losses. See Algorithm~\ref{algo:train_epoch} for further details. \textbf{Right}: Image classification using image similarity for few-shot classification without using labels. Images are encoded using the model learned in the training phase. Then the classifier predicts using the most similar key image for every query image. This allows us to classify images with no label information at test time. See Algorithm~\ref{algo:test_task} for further details. \textit{(Best viewed in color)}}
  \vspace{-5pt}
  \label{fig:pipeline_unified}
\end{figure*}

A different category of approaches constrain the model's representational capacity or hypothesis space. With a smaller hypothesis space to explore, the model is able to quickly train using a few samples. Multitask learning methods~\cite{caruana1997multitask,zhang2017survey_multitask} leverage multiple related tasks by learning them simultaneously. Knowledge is aggregated by sharing~\cite{zhang2018fine,motiian2017few,benaim2018one} or tying~\cite{yan2015multi} parameters together. Embedding learning methods~\cite{jia2014caffe} learn to embed inputs into a lower dimensional space such that similar inputs are embedded closer than dissimilar ones. Matching Networks~\cite{matchNets} and Prototypical Networks~\cite{protoNets} fall into this category: \cite{matchNets} learns train and test-specific embeddings, whereas \cite{protoNets} learns to embed class prototypes. Memory networks~\cite{miller2016key, ramalho2018adaptive} directly store training examples in external memory for use at test time. Choosing appropriate examples~\cite{zhu2018compound, xu2017few} to store is a non-trivial task and directly affects performance. Generative methods~\cite{reed2018few, salakhutdinov2012one} aim to directly learn the underlying data generating probability distribution. Our work does not constrain the model at all in any way, using a completely different approach for achieving our objectives.

A third category of methods guides the search strategy for optimal parameters, by either learning a good initialization or guiding the parameter updates. Approaches such as \cite{caelles2017one, yoo2018efficient,kozerawski2018clear} directly fine-tune parameters learned from a different but related pre-text task. Meta-learning methods~\cite{pmlr-v70-finn17a} aim to learn a good initialization from multiple related tasks, which can be optimized for specific tasks in a few gradient updates~\cite{Bengio97onthe,NaikMetaNets}. Advances in the meta-learning space focus on incorporating task-specific information~\cite{lee2018gradient}, modeling uncertainty~\cite{prob_MAML, bayes_MAML, recast_MAML,rusu2018metalearning}, or improving training strategy~\cite{how_MAML,alpha_MAML}. Finally, \cite{Ravi2017OptimizationAA, andrychowicz2016learning} learn an optimizer that directly guides parameter updates of the task specific learner. In contrast, we do not update any network parameters at test time.

There has been a recent trend towards self-supervision and metric learning approaches for few-shot learning which has motivated our work. Su \textit{et.\ al}~\cite{maji_ssl_boost} and Gidaris \textit{et\ al.}~\cite{gidaris_ssl_boost} use self-supervision as an auxiliary task to improve the representations learnt using supervised pipelines. AmDimNet~\cite{amdim_net} uses a self-supervised pre-training step but, contrary to our work, uses label information at test time to refine parameters using meta-learning. Works such as \cite{wang2019simpleshot,chen2019closerfewshot} have established the effectiveness of simple nearest neighbor classifiers for few-shot tasks. Contrary to our work, these efforts use a supervised training setup. A very recent work~\cite{wvangansbeke2020scan} has goals similar to ours in terms of learning to classify without labels, but does not address few-shot learning.

Most aforementioned methods are computationally intensive~\cite{pmlr-v70-finn17a}, conceptually complex~\cite{MTL}, or simply shift the burden of annotations from the novel to the base classes, often requiring annotated data from multiple tasks. Our current work combines label-free training with extremely simple classifiers, leading to a completely label-free training and testing setup for the few-shot regime.

\subsection{Other Related Perspectives}
\vspace{-3pt}
We briefly review earlier work on metric learning and self-supervised learning, which we leverage to achieve our objectives in this work.

\noindent \textbf{Metric learning} methods, which are used for capturing image similarity, ``learn to compare''. By learning image similarity, a model can use similarity to label instances of novel classes by comparing to known examples. It is also an effective pre-text task for few-shot learning~\cite{Koch2015SiameseNN}. These models learn by conditioning predictions on distance metrics such as cosine similarity~\cite{matchNets}, Euclidean distances~\cite{protoNets}, network based~\cite{sung2018RelationNet}, ridge regression~\cite{bertinetto2018r2d2}, convex optimization based~\cite{meta_opt_net}, or graph neural networks~\cite{garcia2018fewshot}. Regularization techniques such as manifold mixup~\cite{charting_manifold}, combined with supervised pipelines, also improve accuracies.

\noindent \textbf{Self-supervised} methods remove the need for annotated data by using a supervisory signal from the data itself. A number of pretext tasks such as predicting image colorization~\cite{larsson2016autoColor,zhang2016colorful}, image patch positions~\cite{norooziECCV16,doersch2015unsupervised}, and image rotations~\cite{gidaris2018unsupervised} are used in literature but do not address the few-shot problem. Combining self-supervision with supervised approaches~\cite{gidaris_ssl_boost,maji_ssl_boost,amdim_net} has been shown to improve accuracies over few-shot tasks, but these methods are not entirely label-free. Finally,~\cite{chen2020simple, he2019moco, tian2019cmc, ji2019iic} learn contrastive representations by applying simple transforms on input images and predicting image similarity. By learning to predict image similarity in the presence of distortions, the network is able to effectively distill information, making it suitable for quick adaptation on novel classes. In our work, we leverage two recent contrastive approaches: SimCLR~\cite{chen2020simple} and MoCo~\cite{he2019moco} to learn appropriate image representations without labels. We now present our methodology.

\section{Methodology: Few-Shot Learning with No Labels}
\vspace{-3pt}
We now present the details of our training and testing methods. Figure~\ref{fig:pipeline_unified} presents an overview of both these stages. During training, we learn contrastive representations in a self-supervised manner. At test time, we use the learned representations to encode images and leverage image similarity to perform few-shot classification. Thus, we are able to achieve strong baselines without using any labels during training or testing.

Following recent work~\cite{he2019moco, chen2020simple, wang2019simpleshot}, we follow a simple two-phase (train + test) approach. In the training phase, the base network is trained using self-supervised approaches on the base classes alone. During the test phase, the trained network is used to obtain representations, which are provided to very simple classifiers on the few-shot classes. 
We ensure that no label information is available to the network at any time. Even though we have access to $K$ labeled examples per $C$ classes, the network does not access the labels, keeping in line with our label-free setting. Our reasons for using only very simple classifiers during the testing phase are two-fold: firstly, recent literature~\cite{chen2020simple, protoNets, matchNets, chen2019closerfewshot} has demonstrated the competitiveness of simple baselines, and secondly, these classifiers require no labels during test time, allowing us to solely focus on the effectiveness of our training scheme and maintaining the label-free setting.

\subsection{Proposed Training Methodology}
\vspace{-3pt}
Since learning image similarity is a key strategy used in few-shot tasks~\cite{Koch2015SiameseNN}, we focus on learning contrastive representations during our training phase. Unlike \cite{Koch2015SiameseNN}, this allows us to completely ignore the labels. Given an input image, an augmentation module $\mathcal{A}$ applies image transforms to generate two images, the query $q$ and the key $k$. Contrastive learning frameworks maximize similarity between images generated from the same input, and maximize dis-similarity between images which are generated from different base inputs. We then have three steps in our training method, as  visualized in Figure~\ref{fig:pipeline_unified} and outlined in Algorithm~\ref{algo:train_epoch}.

First, a stochastic \textit{data augmentation} module $\mathcal{A}$ generates pairs of image minibatches ($\mathcal{M}_{\mathcal{A}}^q, \mathcal{M}_{\mathcal{A}}^k$) from the input minibatch $\mathcal{M}$ using image transforms. The first minibatch consists of query images $q$, and the second minibatch consists of key images $k$. For every query image $q$, the key that was generated from the same base image is designated as a positive sample $k^+$ and the pair $(q, k^+)$ is designated as a positive pair. The other key images form a set of negative samples $\{k^-\}$ and the pairs $(q, k^-)$ are designated negative pairs. We then leverage the recent success of self-supervised contrastive learning approaches, in particular, SimCLR~\cite{chen2020simple} and MoCo~\cite{he2019moco}, as two variants in our overall strategy. The choice between these two methods affects how negative samples are generated and handled. More details of how these methods are used are presented later in this section.

Secondly, deep neural \textit{encoder networks} $f(\cdot)$, and $g(\cdot)$ encode the pair of augmented image batches, to learn an appropriate image representation. Network $f(\cdot)$ learns the representation $R_q$ of query images, while network $g(\cdot)$ learns the representations $R_k$ of key images. 
We only use the network $f(\cdot)$ for downstream tasks.

Finally, the encoded representations are fed into a \textit{constrastive loss function} $\mathcal{L}$ for the contrastive prediction task. Given the representations of an input query image $R_q$, a positive example $R_{k^+}$, and a set of negative examples $\{R_{k^-}\}$, the contrastive prediction task aims to maximize the similarity of representation between the query and the positive sample, and minimize the similarity of representation between the query and the negative samples. Given a similarity function $s(\mathbf{x}, \mathbf{y})$ the contrastive loss used in our methodology is given by:
\begin{equation}
\begin{split}
  & \mathcal{L}(R_q, R_{k^+}, \{R_{k^-}\}) = \\
  & -\log{\frac{\exp{s(R_q, R_{k^+})/\tau}}{\exp{s(R_q, R_{k^+})/\tau} + \underset{R_{k^-}}{\sum}\exp{s(R_q, R_{k^-})/\tau}}}
\end{split}
\label{eqn_contrastive_loss}
\end{equation}
where $\tau$ is a temperature hyperparameter.

We now present specific details of our training method variants.

\begin{algorithm}[t]
  \SetAlgoLined
  
  \SetKwFunction{ChoosePos}{ChoosePositive}
  \SetKwFunction{ChooseNeg}{ChooseNegative}
  \SetKwFunction{Update}{UpdateParams}
  
  \KwIn{Augmentation Module $\mathcal{A}(\cdot)$}
  \KwIn{Encoders $f(\cdot)$ $g(\cdot)$}
  \KwIn{Constrastive Loss Module $\mathcal{L}(q, k^+, \{k^-\})$}
  \KwData{Training dataset $\mathcal{D}_{tr}$}
  \KwResult{Trained network $f(\cdot)$}
  \BlankLine
  \For{minibatch $\mathcal{M}$ in $\mathcal{D}_{tr}$}{
    \BlankLine
    \tcp{get augmented minibatches}
    $\mathcal{M}_{\mathcal{A}}^q, \mathcal{M}_{\mathcal{A}}^k = \mathcal{A}(\mathcal{M})$\;
    \BlankLine
    \tcp{get representations}
    $\{R_q\} = f(\mathcal{M}_{\mathcal{A}}^q)$\;
    $\{R_k\} = g(\mathcal{M}_{\mathcal{A}}^k)$\;
    \BlankLine
    \For{query $R_q$ in $\{R_q\}$}{
      \tcp{positive key for query}
      $R_{k^+}$ = \ChoosePos{$R_q$, $\{R_k\}$}\;
      \BlankLine
      \tcp{negative keys for query}
      $\{R_{k^-}\}$ = \ChooseNeg{$R_q$, $\{R_k\}$}\;
      \BlankLine
      \tcp{contrastive loss}
      $\mathcal{L}(R_q, R_{k^+}, \{R_{k^-}\})$\;
      \BlankLine
      \tcp{update network parameters}
      \Update{$f$, $g$}\;
    }
  }
  \Return{$f$}
  \caption{Overall Training Methodology}
  \label{algo:train_epoch}
\end{algorithm}

\subsubsection{SimCLR Training}
\vspace{-3pt}
In the SimCLR setting, we start from an input minibatch of $N$ images and generate two augmented batches by performing two different image transforms. SimCLR treats both batches on equal footing with no distinction between \textit{query} and \textit{key} images. Considering pairs with different image order to be distinct, there are $2N$ positive pairs formed by images which have been generated from the same input, and $2N(2N-2)$ negative pairs which are formed by different images. The same network $f(\cdot)$ is used to embed both keys and queries. A cosine similarity function is used in the contrastive loss (Eqn \ref{eqn_contrastive_loss}), $s(\mathbf{x}, \mathbf{y}) = \mathbf{x}^T\mathbf{y} / |\mathbf{x}| |\mathbf{y}|$. This setup is referred to as \textsc{Ours\_S} in our results.

\subsubsection{MoCo Training}
\vspace{-3pt}
The MoCo setting decouples the number of negative samples from the batch size. Once the key and query images have been generated from the input, we formulate the few-shot task as a dictionary lookup problem. The dictionary consists of \textit{key} images, and the unknown image to be looked up is the \textit{query}. The keys are encoded using a momentum encoder, which maintains the set of positive and negative samples per query. The query (non-momentum) encoder is used for downstream few-shot tasks. This setting uses a dot product as the similarity function for contrastive loss $s(\mathbf{x}, \mathbf{y}) = \mathbf{x}^T\mathbf{y}$ and is referred to as \textsc{Ours\_M} in our results.

\begin{algorithm}[t]
\SetAlgoLined

\SetKwData{CCount}{correct}

\KwIn{Trained Encoder $f$}
\KwIn{Classifier $\mathcal{C}_f$}
\KwIn{Similarity Function $s(\mathbf{x}, \mathbf{y})$}
\KwData{$N \times Q$ query images: $\{(q_i, y_{q_i})\}$}
\KwData{$N \times K$ test images: $\{(k_i, y_{k_i})\}$}

\KwResult{Accuracy on task}

$\CCount \leftarrow 0$\;
\ForEach{query image $q_i$}{
  \tcp{return index of most similar key since classifier has no label access}
  $l = \mathcal{C}_f(q_i, \{k_j\})$\;
  \lIf{$y_{q_i} == y_{k_l}$}{$\CCount = \CCount + 1$}
}
\Return{$\CCount / (N \times Q)$}
\caption{Test Phase: $N$-way, $K$-shot Task}
\label{algo:test_task}
\end{algorithm}

\subsection{Inference}
\label{subsec_inference}
\vspace{-3pt}
The testing phase consists of few-shot tasks, following earlier literature~\cite{wang2019simpleshot}. Each $C$-way $K$-shot task consists of $K$ key images, and $Q$ query images from $C$ classes each. Using the representations learnt, the network must classify query images.

Given the set of key images $\{k\}$, a query image $q$ to be classified, and the trained network $f(\cdot)$ from the training phase, our classifier $\mathcal{C}_f$ matches $q$ with its corresponding key image $k_j$ by returning the \textit{index} $j$ of the most similar key image. The matching process is described later in this section. By returning the index of the most similar key image, we are able to classify the query without access to labels. Figure~\ref{fig:pipeline_unified} also presents a visual overview.

Inspired by \cite{chen2019closerfewshot,wang2019simpleshot} we study the use of two different test time classifiers: the 1-Nearest Neighbor classifier (\textsc{1NN}) from SimpleShot~\cite{wang2019simpleshot} and a soft cosine attention kernel (\textsc{Attn}) adapted from Matching Networks~\cite{matchNets}. Our inference methodology is outlined in Algorithm \ref{algo:test_task}. The \textsc{1NN} classifier chooses the key image which minimizes the Euclidean distance between key and the query image under consideration.
\begin{equation}
    \mathcal{C}_f(q, \{k\}) = \argmin_j{|f(q) - f(k_j)|^2}
\end{equation}
The \textsc{Attn} classifier chooses the key image corresponding to each query using an attention mechanism which provides a softmax over the cosine similarities. Unlike Matching Networks~\cite{matchNets}, we take an $\argmax$ instead of a weighted average over the labels of the key image set. This is because our classifier has no access to the probability distribution over the labels, or the number of labels.
\begin{equation}
\begin{split}
    \mathcal{C}_f(q, \{k\}) &= \argmax_j{a_{\{k\}}(q, k_j)} \\
    a_{\{k\}}(q, k_j) &= \frac{\exp{c(f(q), f(k_j))}}{\sum_i\exp{c(f(q), f(k_i))}} \\
    c(\mathbf{x}, \mathbf{y}) &= \frac{\mathbf{x} \cdot \mathbf{y}}{\mathbf{|x|} \cdot \mathbf{|y|}}
\end{split}
\end{equation}

In the multi-shot setting, it is also possible to compute class centroids as representatives for each class and use those for classification. Since computing class centroids requires label information, we present those experiments as part of our ablation studies in Section \ref{subsection:ablation}.
\begin{table*}
  \centering
  \begin{tabular}{c | l c c | c c c c}
    \hline
    \multirow{2}{*}{\textbf{Pipeline}} & \multirow{2}{*}{\textbf{Approach}} & \multicolumn{2}{c|}{\textbf{Setting}} & \multicolumn{4}{c}{\textbf{Labels Used}} \\
    & & \textbf{1-shot} & \textbf{5-shot} & \textbf{Total} & \textbf{Train} & \textbf{Test} & \textbf{Validation} \\
    \hline
    \multirow{10}{*}{Supervised} & MAML~\cite{pmlr-v70-finn17a} & 49.6 $\pm$ 0.9 & 65.7 $\pm$ 0.7 & 50,400 & \cmark & \cmark & \xmark \\
    & CloserLook~\cite{chen2019closerfewshot} & 51.8 $\pm$ 0.7 & 75.6 $\pm$ 0.6 & 50,400 & \cmark & \cmark & \xmark \\
    & RelationNet~\cite{sung2018RelationNet} & 52.4 $\pm$ 0.8 & 69.8 $\pm$ 0.6 & 50,400 & \cmark & \cmark & \xmark \\
    & MatchingNet~\cite{matchNets} & 52.9 $\pm$ 0.8 & 68.8 $\pm$ 0.6 & 50,400 & \cmark & \cmark & \xmark \\
    & ProtoNet~\cite{protoNets} &54.1 $\pm$ 0.8 & 73.6 $\pm$ 0.6 & 50,400 & \cmark & \cmark & \xmark \\
    & Gidaris \emph{et al.}~\cite{gidaris_forgetting} & 55.4 $\pm$ 0.8 & 70.1 $\pm$ 0.6 & 50,400 & \cmark & \cmark & \xmark \\
    & TADAM~\cite{tadam} & 58.5 $\pm$ 0.3 & 76.7 $\pm$ 0.3 & 50,400 & \cmark & \cmark & \xmark \\
    & SimpleShot~\cite{wang2019simpleshot} & 62.8 $\pm$ 0.2 & 80.0 $\pm$ 0.1 & 38,400 & \cmark & \xmark & \xmark \\
    
    & Tian \emph{et al.}~\cite{tian2020rethinking} & 64.8 $\pm$ 0.6 & 82.1 $\pm$ 0.4 & 50,400 & \cmark & \cmark & \xmark \\
    
    & S2M2~\cite{charting_manifold} & \textbf{64.9 $\pm$ 0.2} & \textbf{83.2 $\pm$ 0.1} & 50,400 & \cmark & \cmark & \xmark \\
    \hline
    \multirow{3}{*}{Unsupervised} & Antoniou \emph{et al.}~\cite{antoniou2019assume}$^1$ & 33.30 & 49.18 & 21,600 & \xmark & \cmark & \cmark \\
    
    & Wu \emph{et al.}~\cite{wu2018unsupervised}$^2$ & 32.4 $\pm$ 0.1 & 39.7 $\pm$ 0.1 & 0 & \xmark & \xmark & \xmark \\
    
    & Ours & \textbf{50.1 $\pm$ 0.2} & \textbf{60.1 $\pm$ 0.2} & 0 & \xmark & \xmark & \xmark \\

    \hline
  \end{tabular}
  \caption{Average accuracy (in \%) on the miniImageNet dataset. $^1$Results from~\cite{antoniou2019assume}, which did not report confidence intervals. $^2$Results from our experiments adapting the published training code from~\cite{wu2018unsupervised}. \textsc{Ours} was implemented here using \textsc{Ours\_S} pipeline and \textsc{Attn} classifier. See Table~\ref{tab:ablation} for more pipeline and classifier variants.}
  \label{tab:miniImgNet}
\end{table*}

\begin{table*}
  \centering
  \begin{tabular}{c | l c c | c c c c}
    \hline
    \multirow{2}{*}{\textbf{Pipeline}} & \multirow{2}{*}{\textbf{Approach}} & \multicolumn{2}{c|}{\textbf{Setting}} & \multicolumn{4}{c}{\textbf{Labels Used}} \\
    & & \textbf{1-shot} & \textbf{5-shot} & \textbf{Total} & \textbf{Train} & \textbf{Test} & \textbf{Validation} \\
    \hline
    \multirow{7}{*}{Supervised} & MAML~\cite{pmlr-v70-finn17a}$^1$ & 58.9 $\pm$ 1.9 & 71.5 $\pm$ 1.0 & 48,000 & \cmark & \cmark & \xmark \\
    & RelationNet~\cite{sung2018RelationNet}$^1$ & 55.0 $\pm$ 1.0 & 69.3 $\pm$ 0.8 & 48,000 & \cmark & \cmark & \xmark \\
    & ProtoNet~\cite{protoNets}$^1$ & 55.5 $\pm$ 0.7 & 72.0 $\pm$ 0.6 & 48,000 & \cmark & \cmark & \xmark \\
    & R2D2~\cite{bertinetto2018r2d2}$^1$ & 65.3 $\pm$ 0.2 & 79.4 $\pm$ 0.1 & 48,000 & \cmark & \cmark & \xmark \\
    & MetaOptNet~\cite{meta_opt_net} & 72.8 $\pm$ 0.7 & 85.0 $\pm$ 0.5 &60,000 & \cmark & \cmark & \cmark \\
    & Tian \emph{et al.}~\cite{tian2020rethinking} & 73.9 $\pm$ 0.8 & 86.9 $\pm$ 0.5 & 48,000 & \cmark & \cmark & \xmark \\
    & S2M2~\cite{charting_manifold} & \textbf{74.8 $\pm$ 0.2} & \textbf{87.5 $\pm$ 0.1} & 48,000 & \cmark & \cmark & \xmark \\
    \hline
    \multirow{2}{*}{Unsupervised} & Wu \emph{et al.}~\cite{wu2018unsupervised}$^2$ & 27.1 $\pm$ 0.1 & 31.3 $\pm$ 0.1 & 0 & \xmark & \xmark & \xmark \\
    
    & Ours & \textbf{53.0 $\pm$ 0.2} & \textbf{62.5 $\pm$ 0.2} & 0 & \xmark & \xmark & \xmark \\

    \hline
  \end{tabular}
  \caption{Average accuracy (in \%) on the CIFAR100FS dataset. $^1$Results from~\cite{meta_opt_net}. $^2$Results from our experiments adapting the published training code from~\cite{wu2018unsupervised}. \textsc{Ours} was implemented here using \textsc{Ours\_S} pipeline and \textsc{Attn} classifier. See Table~\ref{tab:ablation} for more pipeline and classifier variants.}
  \vspace{-6pt}
  \label{tab:CIFAR100FS}
\end{table*}

\begin{table*}
  \centering
  \begin{tabular}{c| l c c | c c c c}
    \hline
    \multirow{2}{*}{\textbf{Pipeline}} & \multirow{2}{*}{\textbf{Approach}} & \multicolumn{2}{c|}{\textbf{Setting}} & \multicolumn{4}{c}{\textbf{Labels Used}} \\
    & & \textbf{1-shot} & \textbf{5-shot} & \textbf{Total} & \textbf{Train} & \textbf{Test} & \textbf{Validation} \\
    \hline
    \multirow{5}{*}{Supervised} & ProtoNet~\cite{protoNets}$^1$& 35.3 $\pm$ 0.6 & 48.6 $\pm$ 0.6 & 48,000 & \cmark & \cmark & \xmark \\
    & TADAM~\cite{tadam}$^1$ & 40.1 $\pm$ 0.4 & 56.1 $\pm$ 0.4 & 48,000 & \cmark & \cmark & \xmark \\
    & MTL~\cite{MTL} & 45.1 $\pm$ 1.8 & 57.6 $\pm$ 0.9 & 60,000 & \cmark & \cmark & \cmark \\
    & MetaOptNet~\cite{meta_opt_net} & \textbf{47.2 $\pm$ 0.6} & \textbf{62.5 $\pm$ 0.6} & 60,000 & \cmark & \cmark & \cmark \\
    & Tian \emph{et al.}~\cite{tian2020rethinking} & 44.6 $\pm$ 0.7 & 60.9 $\pm$ 0.6& 48,000 & \cmark & \cmark & \xmark \\
    \hline

    \multirow{2}{*}{Unsupervised} & Wu \emph{et al.}~\cite{wu2018unsupervised}$^2$ & 27.4 $\pm$ 0.1 & 32.4 $\pm$ 0.1 & 0 & \xmark & \xmark & \xmark \\
    
    & Ours & \textbf{37.1 $\pm$ 0.2} & \textbf{43.4 $\pm$ 0.2} & 0 & \xmark & \xmark & \xmark \\

    \hline
  \end{tabular}
  \caption{Avg accuracy (in \%) on FC100 dataset. $^1$Results from~\cite{meta_opt_net}. $^2$Results from our experiments adapting published training code from~\cite{wu2018unsupervised}. \textsc{Ours} was implemented here using \textsc{Ours\_S} pipeline and \textsc{Attn} classifier. See Table~\ref{tab:ablation} for more pipeline and classifier variants.}
  \label{tab:FC100}
\end{table*}

\begin{table*}
  \centering
  \begin{tabular}{cc|cc|cc|cc}
    \hline
    \multirow{2}{*}{\textbf{Training}} & \multirow{2}{*}{\textbf{Testing}} & \multicolumn{2}{c|}{\textbf{miniImagenet}} & \multicolumn{2}{c|}{\textbf{CIFAR100}} & \multicolumn{2}{c}{\textbf{FC100}} \\
    & & \textbf{1-shot} & \textbf{5-shot} & \textbf{1-shot} & \textbf{5-shot} & \textbf{1-shot} & \textbf{5-shot} \\
    \hline
    \multirow{4}{*}{Ours\_S} & 1NN & 48.7 $\pm$ 0.2 & 59.0 $\pm$ 0.2 & 52.0 $\pm$ 0.2 & 61.7 $\pm$ 0.2 & 36.0 $\pm$ 0.2 & 42.6 $\pm$ 0.2 \\
    & Attn & 50.1 $\pm$ 0.2 & 60.1 $\pm$ 0.2 & \textbf{53.0 $\pm$ 0.2} & 62.5 $\pm$ 0.2 & \textbf{37.1 $\pm$ 0.2} & 43.4 $\pm$ 0.2 \\
    & 1NN\_centroid & - & 64.6 $\pm$ 0.2 & - & \textbf{65.8 $\pm$ 0.2} & - & \textbf{47.2 $\pm$ 0.2} \\
    & Attn\_centroid & - & 63.6 $\pm$ 0.2 & - & 63.8 $\pm$ 0.2 & - & 46.0 $\pm$ 0.1 \\ \hline
    
    \multirow{4}{*}{Ours\_M} & 1NN & 29.7 $\pm$ 0.1 & 39.0 $\pm$ 0.1 & 26.7 $\pm$ 0.1 & 31.1 $\pm$ 0.2 & 28.1 $\pm$ 0.1 & 33.2 $\pm$ 0.2 \\
    
    & Attn & 36.0 $\pm$ 0.2 & 45.2 $\pm$ 0.1 & 27.9 $\pm$ 0.1 & 32.3 $\pm$ 0.2 & 30.4 $\pm$ 0.1 & 35.2 $\pm$ 0.2 \\
    
    & 1NN\_centroid & - & 45.1 $\pm$ 0.2 & - & 32.4 $\pm$ 0.2 & - & 34.9 $\pm$ 0.2 \\
    
    & Attn\_centroid & - & 48.4 $\pm$ 0.2 & - & 32.4 $\pm$ 0.1 & - & 35.6 $\pm$ 0.2 \\ \hline
    
    \multirow{4}{*}{Ours\_SF} & 1NN & 46.8 $\pm$ 0.2 & 61.2 $\pm$ 0.1 & 42.5 $\pm$ 0.2 & 56.1 $\pm$ 0.2 & 30.9 $\pm$ 0.1 & 39.0 $\pm$ 0.2 \\
    & Attn & \textbf{55.6 $\pm$ 0.2} & 68.2 $\pm$ 0.2 & 49.3 $\pm$ 0.2 & 61.6 $\pm$ 0.2 & 33.4 $\pm$ 0.2 & 41.3 $\pm$ 0.2 \\
    & 1NN\_centroid & - & \textbf{72.4 $\pm$ 0.1} & - & 65.7 $\pm$ 0.2 & - & 44.6 $\pm$ 0.2 \\
    & Attn\_centroid & - & 72.1 $\pm$ 0.1 & - & 64.9 $\pm$ 0.2 & - & 43.5 $\pm$ 0.2 \\
    \hline
  \end{tabular}
  \caption{An ablation study of multiple classifiers on various backbones. Average accuracy and 95\% confidence intervals are reported over 10,000 rounds. The \texttt{\_centroid} classifiers use class labels to compute the centroids per class. Best results per dataset and few-shot task are in \textbf{bold}.}
  \label{tab:ablation}
\end{table*}

\section{Experiments}
\vspace{-3pt}
We study the proposed methodology on various few-shot classification benchmarks. All our training and testing methods on each of these benchmark datasets are label-free. Classification is done via image similarity using non-parametric classifiers, as described above. 

\subsection{Experimental Setup}
\vspace{-3pt}
We describe the experimental setup in this section: Section \ref{subsubsection:datasests} describes the datasets~\cite{matchNets, cifar100_base, tadam} which we use for few-shot image classification; Section \ref{subsubsection:evaluation} presents our evaluation strategy for reporting results; and Section \ref{subsubsection:implementation} presents the hyperparameters for our models and implementation details for reproducibility.

\subsubsection{Datasets}
\label{subsubsection:datasests}
\vspace{-3pt}
We perform our experiments on three popularly used few-shot image classification benchmark datasets, each of which is described below.

The \textbf{\emph{mini}ImageNet} dataset~\cite{matchNets} is a subset of ImageNet~\cite{imagenet_cvpr09} and is a common few-shot learning benchmark. The dataset contains 100 classes and 600 examples per class. Following~\cite{Ravi2017OptimizationAA}, we split the dataset to have 64 base classes, 16 validation classes, and 20 novel classes. Following~\cite{matchNets}, we resize the images to 84 $\times$ 84 pixels via rescaling and center cropping.

We also perform experiments on a subset of the CIFAR-100~\cite{cifar100_base} dataset, as in \cite{tadam}. This dataset consists of 100 image classes in total with each class having 600 images of size 32 $\times$ 32 pixels. Following the setup in~\cite{tadam}, we split the classes into 60 base, 20 validation, and 20 novel classes for few-shot learning. This dataset is referred to as \textbf{CIFAR-100FS} in our experiments.

We also use the \textbf{FC100}~\cite{tadam} (FewShot CIFAR100) dataset for our experiments. The 100 classes of the CIFAR-100~\cite{cifar100_base} dataset are grouped into 20 superclasses to minimize information overlap. The train split contains 60 classes belonging to 12 superclasses, the validation and test splits contain 20 classes belonging to 5 superclasses each. $Q$ (as in Section \ref{subsec_inference}) is chosen to be 15 across all datasets.

\subsubsection{Evaluation Protocol}
\label{subsubsection:evaluation}
\vspace{-3pt}
We follow a standard evaluation protocol following earlier literature in the field~\cite{rusu2018metalearning, wang2019simpleshot}. The classifier is presented with 10,000 tasks and average accuracy is reported. 
Given a test set consisting of $C$ novel classes, we generate a $N$-way $K$-shot task as follows. $N$ classes are uniformly sampled from the set of $C$ classes without replacement. From each class, $K$ key and $Q=15$ query images are uniformly sampled without replacement. The classifier is presented with the key images and then used to classify the query images.
Following prior work~\cite{wang2019simpleshot}, we focus on 5-way 1-shot and 5-way 5-shot benchmarks. 

\subsubsection{Models and Implementation Details}
\label{subsubsection:implementation}
\vspace{-3pt}

All experiments use a ResNet-50~\cite{He2015} backbone. SimCLR~\cite{chen2020simple} pre-training is done for  500 epochs with a learning rate of 0.1, nesterov momentum of 0.9, and weight decay of 0.0001 on the respective datasets. Data augmentations of RandomResizedCrop and ColorDistortion were found to achieve the best results. The augmentations use default hyperparameters from \cite{chen2020simple}. More choices of hyperparameters and corresponding performance are presented in the supplementary section, due to space constraints.

MoCo~\cite{he2019moco} pre-training is done for 800 epochs over the respective training sets using the default parameters from MoCo-v2~\cite{chen2020mocov2}. Downstream tasks use the query (non-momentum) encoder network.

\subsection{Results}
\vspace{-3pt}
Tables \ref{tab:miniImgNet}, \ref{tab:CIFAR100FS} and \ref{tab:FC100} present our results on the \textit{mini}ImageNet, CIFAR100FS and FC100 datasets respectively. For a more comprehensive comparison, we also adapt the work presented in Wu \emph{et al.}~\cite{wu2018unsupervised} to include another unsupervised method in these results.
Accuracies are averaged over 10,000 tasks and reported with 95\% confidence intervals. 
Note that we report the number of labels used by each method in each of the above tables. The number of labels used by the methods are computed as follows: if the network trains by performing gradient updates over the training labels, we count the labels in the training set; if the network fine-tunes over the test labels or uses test labels to compute class representations, we count the labels in the test set; if the network uses training and validation data to report results, we count training and validation labels. Unless otherwise specified in the respective works, we assume that the validation set has not been used to publish results, and that the train and test pipelines are the same.
\begin{figure*}
  \centering
  \includegraphics[width=0.9\linewidth]{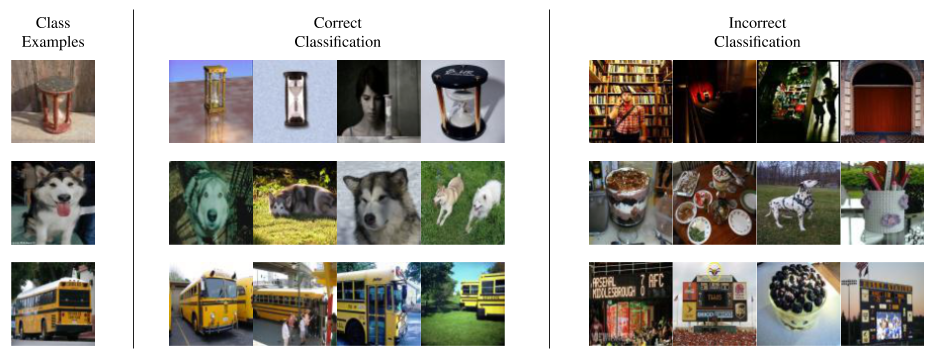}
  \vspace{-5pt}
  \caption{Visualizing a few examples from the miniImagenet test set using the \textsc{Ours\_S} pipeline. \textbf{Far Left:} One labelled example visualized per class. \textbf{Middle:} Few correctly classified examples from the test set. \textbf{Right:} Mis-classified examples. Similarity in texture and coarse object category are contributing factors for mis-classification.}
  \vspace{-5pt}
  \label{fig:mosaic}
\end{figure*}

We observe that our method achieves strong baselines on the benchmarks without using any label information at all, as can be seen in the comparison with Wu \emph{et al.}~\cite{wu2018unsupervised} which operates in the same setting. These are the only two methods across all these results that use \textbf{zero} label information. All other methods are provided for comparison purposes, and use label information for backpropagation, for generating positive and negative sample pairs, for fine-tuning at test time, or for calculating class representatives at test time. The label information used in these supervised methods are of the order of tens of thousands, which can be very expensive depending on a particular domain. Our methodology seeks to provide a pathway to solving problems in such settings with no annotation cost whatsoever.

The 5-way 5-shot tasks are relatively simpler, and our results reflect that across all settings. The best results are achieved over the challenging miniImageNet dataset, followed by CIFAR100FS and FC100 datasets. This is to be expected as FC100 is a coarse-grained classification task and is specifically constructed to have dissimilar classes.

In Section \ref{subsec_inference}, we proposed the use of two test-time classifiers: \textsc{1NN} and \textsc{Attn}. We report ablation studies on their performances in Table \ref{tab:ablation}. While the \textsc{1NN} classifier achieves strong baselines (in line with previous work~\cite{chen2020simple}), the \textsc{Attn} classifier consistently improves accuracies by 2-10\%, with more impressive gains in the multi-shot setting.

\subsection{Ablation Studies}
\label{subsection:ablation}
\vspace{-5pt}
In this section, we explore different variations of our pipeline and compare the performance across datasets. Table \ref{tab:ablation} presents the results.

\noindent \textbf{Leveraging Clustering Methods:} Since we leverage image similarity for few-shot classification at test time, a simple way to increase accuracy is to ensure that the representations  of the images from the same class lie close to each other in the embedding space. Hence, we study the utility of the SelfLabel~\cite{asano2020self_label} (\textsc{Ours\_SF}) method as an additional pre-training framework. This work learns to simultaneously cluster images and learn representations without labels, and hence is a good candidate method to integrate in our framework. As shown in Table \ref{tab:ablation}, this improves performance on mini-ImageNet by a significant amount. However, \textsc{Ours\_SF} approach does not yield benefits on other datasets. Since we use image similarity for classification, improving the cluster quality via self-supervision could be a promising direction for future work.

\noindent \textbf{What if we had labels?}  To investigate the effect of introducing labels at test time, we introduce \textsc{centroid} versions of our classifiers: 1 Nearest Neighbours Centroid (\textsc{1NN\_centroid}), and Soft Cosine Attention Centroid (\textsc{Attn\_centroid}), in the multi-shot setting. Following \cite{protoNets, wang2019simpleshot, matchNets}, the \textsc{centroid} versions of these classifiers compute class representatives as the centroids of the key images provided at test time. Few-shot classification is then done by comparing each query image against each class centroid, essentially treating the class representative (or exemplar) as the new key image for that class. 
Using label information to compute class centroids increases performance by 2-4\%.

\noindent \textbf{Qualitative Analysis:}  Figure~\ref{fig:mosaic} presents a few qualitative examples from our results on the \textbf{\textit{mini}ImageNet} dataset using our \textsc{Ours\_S} pipeline. In the second row, we observe that the network fails on a fine-grained classification task. It classifies a \textsc{dalmatian} image (black and white polka-dotted dog) as a \textsc{husky}. Since both categories are dog breeds, they are closely related and poses a difficult few-shot problem. However, when the classes are coarse-grained and fairly well-separated, our method shows that one can achieve reasonable performance with no label information at all.

\section{Conclusion}
\vspace{-5pt}
We present a new framework for few-shot classification without any label information. This is a more challenging task than existing work which uses label information at various points during training or inference. By learning contrastive representations using self supervision, we achieve competitive baselines while using \textbf{zero} labels, which is a few \textbf{10,000 times} fewer labels than existing work.
In our ablation studies, we analyze the effect of using a different training setup attuned towards clustering, and also observe results when label information is added to the pipeline, in a limited way. Since we perform few-shot classification using image similarity, clustering quality directly impacts classification accuracy and improving it is a promising direction for future work. Our ablations also show that using limited label information to compute class representatives at test time is beneficial. Our objective in this work was to not achieve state-of-the art performance, but to show that one can obtain reasonable performance on few-shot classification with zero labels. We believe this work is an important first step towards label-free few-shot learning methods.

\clearpage

{\small
\bibliographystyle{ieee_fullname}
\bibliography{egbib}
}

\end{document}


\title{Supplementary Material:\\Few-Shot Learning with No Labels}


\maketitle


\renewcommand{\citenumfont}[1]{A#1} 
\renewcommand{\bibnumfmt}[1]{[A#1]} 
\appendix

In this supplementary material, we present additional experiments and investigations about our few-shot learning method without any labels. We first investigate the impact of choosing appropriate augmentation transforms for few shot learning. We show that the choice of augmentation transform has a significant impact on the quality of representations learned. Next, we investigate the effect of the dimension of the learned representations on the accuracy. Since our classifiers perform classification via image similarity, the dimension of the representations may affect the final results.

\begin{table*}
  \centering
  \begin{tabular}{c c|cc|cc|cc}
    \hline
    \multirow{2}{*}{\textbf{Transforms}} & \multirow{2}{*}{\textbf{Classifier}} & \multicolumn{2}{c|}{\textbf{miniImageNet}} & \multicolumn{2}{c|}{\textbf{CIFAR100FS}} & \multicolumn{2}{c}{\textbf{FC100}} \\
    & & \textbf{1-shot} & \textbf{5-shot} & \textbf{1-shot} & \textbf{5-shot} & \textbf{1-shot} & \textbf{5-shot} \\
    \hline
    \multirow{2}{*}{Crop+Blur} & 1NN & 33.47 $\pm$ 0.16 & 41.92 $\pm$ 0.16 & 36.17 $\pm$ 0.17 & 47.43 $\pm$ 0.18 & 31.91 $\pm$ 0.15 & 41.14 $\pm$ 0.16 \\
    & Attn & 34.79 $\pm$ 0.17 & 42.65 $\pm$ 0.16 & 37.97 $\pm$ 0.18 & 47.79 $\pm$ 0.18 & 32.80 $\pm$ 0.16 & 41.30 $\pm$ 0.16 \\
    \hline
    \multirow{2}{*}{Crop+Distort} & 1NN & 48.65 $\pm$ 0.20 & 58.98 $\pm$ 0.18 & 51.96 $\pm$ 0.24 & 61.68 $\pm$ 0.20 & 36.00 $\pm$ 0.18 & 42.54 $\pm$ 0.17 \\
    & Attn & \textbf{50.13 $\pm$ 0.21} & \textbf{60.09 $\pm$ 0.18} & \textbf{52.90 $\pm$ 0.24} & \textbf{62.46 $\pm$ 0.21} & \textbf{37.09 $\pm$ 0.19} & \textbf{43.39 $\pm$ 0.18} \\
    \hline
    \multirow{2}{*}{Distort+Blur} & 1NN & 21.96 $\pm$ 0.09 & 24.68 $\pm$ 0.10 & 24.90 $\pm$ 0.12 & 30.23 $\pm$ 0.15 & 22.08 $\pm$ 0.08 & 25.08 $\pm$ 0.10 \\
    & Attn & 22.40 $\pm$ 0.10 & 25.77 $\pm$ 0.10 & 25.55 $\pm$ 0.13 & 31.10 $\pm$ 0.15 & 22.92 $\pm$ 0.10 & 26.93 $\pm$ 0.11 \\
    \hline \\
  \end{tabular}
  \caption{Results of our experiments with different combinations of data transforms. \textbf{Crop} refers to RandomResizedCrop. \textbf{Blur} refers to GaussianBlur. \textbf{Distort} refers to ColorDistortion. Representations learned using the Ours\_S pipeline on the respective datasets. Accuracies averaged over 10,000 tasks and 95\% confidence intervals reported. Best results in each setting in \textbf{bold}.}
  \label{tab:augmentation}
\end{table*}

\begin{table*}
  \centering
  \begin{tabular}{c c|cc|cc|cc}
    \hline
    \multirow{2}{*}{\textbf{Output Dim}} & \multirow{2}{*}{\textbf{Classifier}} & \multicolumn{2}{c|}{\textbf{miniImageNet}} & \multicolumn{2}{c|}{\textbf{CIFAR100FS}} & \multicolumn{2}{c}{\textbf{FC100}} \\
    & & \textbf{1-shot} & \textbf{5-shot} & \textbf{1-shot} & \textbf{5-shot} & \textbf{1-shot} & \textbf{5-shot} \\
    \hline
    2048 & 1NN & 46.79 $\pm$ 0.20 & 61.17 $\pm$ 0.18 & 42.53 $\pm$ 0.19 & 56.11 $\pm$ 0.18 & 30.93 $\pm$ 0.14 & 38.92 $\pm$ 0.16 \\
    (no PCA) & Attn & \textbf{55.54 $\pm$ 0.20} & \textbf{68.22 $\pm$ 0.16} & \textbf{49.33 $\pm$ 0.20} & \textbf{61.55 $\pm$ 0.17} & \textbf{33.35 $\pm$ 0.15} & \textbf{41.26 $\pm$ 0.16} \\
    \hline
    \multirow{2}{*}{1024} & 1NN & 46.84 $\pm$ 0.20 & 61.24 $\pm$ 0.18 & 42.47 $\pm$ 0.18 & 55.85 $\pm$ 0.17 & 30.97 $\pm$ 0.14 & 38.90 $\pm$ 0.15 \\
    & Attn & \textbf{55.60 $\pm$ 0.20} & \textbf{68.50 $\pm$ 0.17} & \textbf{49.43 $\pm$ 0.19} & \textbf{61.55 $\pm$ 0.17} & \textbf{33.30 $\pm$ 0.16} & \textbf{41.51 $\pm$ 0.15} \\
    \hline
    \multirow{2}{*}{512} & 1NN & 46.72 $\pm$ 0.20 & 61.14 $\pm$ 0.18 & 42.62 $\pm$ 0.19 & 56.07 $\pm$ 0.18 & 30.92 $\pm$ 0.14 & 38.92 $\pm$ 0.16 \\
    & Attn & \textbf{55.51 $\pm$ 0.20} & \textbf{68.15 $\pm$ 0.17} & \textbf{49.47 $\pm$ 0.20} & \textbf{61.66 $\pm$ 0.17} & \textbf{33.40 $\pm$ 0.16} & \textbf{41.45 $\pm$ 0.16} \\
    \hline
    \multirow{2}{*}{256} & 1NN & 46.97 $\pm$ 0.20 & 61.22 $\pm$ 0.17 & 42.77 $\pm$ 0.19 & 56.15 $\pm$ 0.17 & 31.07 $\pm$ 0.14 & 38.86 $\pm$ 0.15 \\
    & Attn & \textbf{55.59 $\pm$ 0.20} & \textbf{68.32 $\pm$ 0.17} & \textbf{49.22 $\pm$ 0.20} & \textbf{61.52 $\pm$ 0.17} & \textbf{33.42 $\pm$ 0.16} & \textbf{41.30 $\pm$ 0.16} \\
    \hline
    \multirow{2}{*}{128} & 1NN & 47.85 $\pm$ 0.20 & 61.81 $\pm$ 0.18 & 43.15 $\pm$ 0.19 & 56.58 $\pm$ 0.17 & 31.00 $\pm$ 0.15 & 38.85 $\pm$ 0.15 \\
    & Attn & \textbf{55.21 $\pm$ 0.20} & \textbf{67.77 $\pm$ 0.17} & \textbf{49.07 $\pm$ 0.20} & \textbf{61.42 $\pm$ 0.17} & \textbf{33.02 $\pm$ 0.16} & \textbf{40.91 $\pm$ 0.15} \\
    \hline \\
  \end{tabular}
  \caption{Results of our experiments with different embedding dimensions, using the Ours\_SF pipeline. After learning network weights during training, we embed the training set and perform PCA to reduce the dimensionality of the representations. At test time we use the same learned transformation matrix to reduce the dimensionality of the test set. Best results for each dimension in \textbf{bold}. Accuracies averaged over 10,000 tasks and 95\% confidence intervals are reported.}
  \label{tab:dimension}
\end{table*}

\section{Data Augmentation}
We learn contrastive representations by augmenting input batches and maximizing similarity between the representations of images generated from the same input. Following \cite{chen2020simple}, we experiment with three candidate data augmentation transforms: \textsc{RandomResizedCrop}, \textsc{GaussianBlur}, and \textsc{ColorDistortion}. In a single training run, we need to use only two of these. Our results are presented in Table~\ref{tab:augmentation}. Before presenting our results, we describe the various data transform techniques in pseudocode for clarity.

For the RandomResizedCrop transform, in addition to randomly cropping and resizing to original size, we also apply a random horizontal flip with probability 0.5.
\begin{verbatim}
from torchvision.transforms import (
  Compose, RandomResizedCrop,
  RandomHorizontalFlip, Normalize)

def RandomResizedCrop(options):
  return Compose([
    RandomResizedCrop(
      size=options.image_size),
    RandomHorizontalFlip(0.5),
    Normalize(options.image_mean,
              options.image_std),
  ])
\end{verbatim}

For the ColorDistortion transform, we first apply the color jitter transform over the 4 channels of the image (RGBA) using a jitter strength hyperparameter, with a probability of 0.8, followed by converting the image to grayscale with probability 0.2.
\begin{verbatim}
def ColorDistortion(options):
  s = options.jitter_strength
  return Compose([
    RandomApply([
      ColorJitter(0.8*s, 0.8*s,
                  0.8*s, 0.2*s)
    ], p=0.8),
    RandomGrayscale(p=0.2),
    Normalize(options.image_mean,
              options.image_std),
  ])
\end{verbatim}

For the GaussianBlur transform, we apply Gaussian Blur with a probability of 0.5. The kernel size is 10\% of the input image size, with a standard deviation chosen randomly between 0.1 to 2.0.
\begin{verbatim}
def GaussianBlur(options):
  kernel_size = options.image_size / 10
  return Compose([
    RandomApply([
      GaussianSmoothing(
        channels=options.image_channels,
        kernel_size=kernel_size,
        sigma=random.uniform(0.1, 2.0),
        dim=2)
    ], p=0.5),
    Normalize(options.image_mean,
              options.image_std),
  ])
\end{verbatim}

Our results are presented in Table~\ref{tab:augmentation}. Best results are achieved using the combination of \textsc{RandomResizedCrop} and \textsc{ColorDistortion} which outperform the other combinations by $\sim$ 15-20\% across each dataset and setting. Such a drastic improvement in accuracy indicates that experimenting with carefully crafted data augmentation transforms is a viable direction for future work and improvements.

\section{Dimensionality Reduction}
At test time, we classify images using image similarity. Since the similarity is directly affected by the distance between representations in the embedded space, we also experiment with reducing the dimension of our learned representations using Principal Component Analysis (PCA). For these experiments, we choose the training method attuned towards clustering, \textsc{Ours\_SF}~\cite{asano2020self_label}. We choose this pipeline since it trains by performing simultaneous clustering and labeling, hence this pipeline is most likely to be affected by the embedding dimension. Without PCA, the original dimension of our embeddings is 2048.

Once our backbone encoder network is trained, we use the learned weights to embed our entire training set and perform PCA. At test time, we use the same transformation matrix to reduce the dimension of our test set. We experiment with various output dimension sizes and present our results in Table~\ref{tab:dimension}.
From the tables, we can see that the reducing the dimension using PCA does not affect the accuracies much. The quality of the learned representations is not affected by a simple linear transform. More complex dimensionality reduction techniques can be attempted as future work to study this further.

{\small
\bibliographystyle{ieee_fullname}
\bibliography{egbib}
}
\clearpage
